\documentclass{article}
\usepackage{spconf,amsmath,graphicx,hyperref}
\usepackage{booktabs}

\title{Scaling Spoken Language Models with Syllabic Speech Tokenization}
\name{Nicholas Lee$^{1}$\qquad  Cheol Jun Cho$^{1}$\qquad Alan W Black$^{2}$\qquad  Gopala K. Anumanchipalli$^1$}

\address{$^1$UC Berkeley\qquad $^2$CMU}

\begin{document}
%
\maketitle
\begin{abstract}
Spoken language models (SLMs) typically discretize speech into high-frame-rate tokens extracted from SSL speech models. 
As the most successful LMs are based on the Transformer architecture, processing these long token streams with self-attention is expensive, as attention scales quadratically with sequence length. A recent SSL work introduces acoustic tokenization of speech at the syllable level, which is more interpretable and potentially more scalable with significant compression in token lengths (4-5 Hz). Yet, their value for spoken language modeling is not yet fully explored. 
We present the first systematic study of syllabic tokenization for spoken language modeling, evaluating models on a suite of SLU benchmarks while varying training data scale. 
Syllabic tokens can match or surpass the previous high-frame rate tokens while significantly cutting training and inference costs, achieving more than a 2× reduction in training time and a 5× reduction in FLOPs. Our findings highlight syllable-level language modeling as a promising path to efficient long-context spoken language models.

\end{abstract}
\section{Introduction}
\label{sec:intro}

\begin{figure}[]
  \centering
  \includegraphics[width=0.5\textwidth]{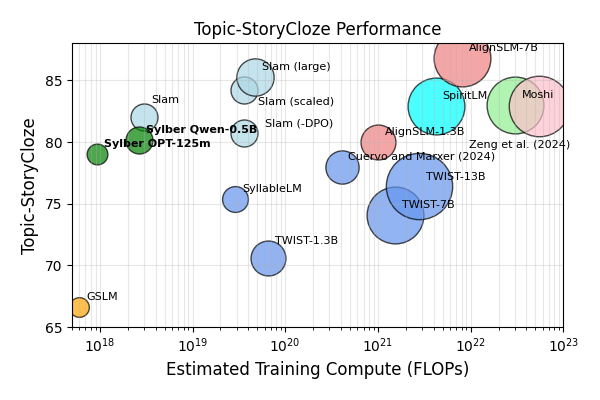}
  \caption{%
    Comparing tSC performance
of different SLMs as a function of training compute, adapted from \cite{maimon2025slamming}.
Sylber based models are shown in dark green on the upper left side.
  }
  \label{fig:tsc_comparison}
  \vspace{-1.5em}
\end{figure}

Speech Language Models (SLMs) have gained popularity in both industry and academia, aiming to transfer the recent success in LMs to the speech domain.
Recent SLMs \cite{ding2025kimi,defossez2024moshi} have been pretrained on millions of hours of data and show promise to potentially envelop and unify separate fields in Speech and Audio such as ASR and TTS; similar to how LLMs unified separate fields of NLP.
Furthermore, they have become a strong foundation for fully audio-based spoken chat systems, offering advantages over prior cascaded ASR–LLM–TTS pipelines.

SLMs typically tokenize speech in two predominant ways.
One way is to use acoustic tokenizers \cite{defossez2022high} which are trained to reconstruct speech and audio, while the other way is to discretize and derive representations from pretrained SSL speech models. \cite{hsu2021hubert, chen2022wavlm}.

Unfortunately, these tokenizers have a central flaw in that they can have very long sequence lengths, typically with a sampling rate of 25-75 Hz.
Since the transformer is the predominant neural network architecture for modeling, the quadratic complexity of attention limits the amount of long-context data that these models can be trained on.
Multiple recent SLM papers \cite{defossez2024moshi, ju2025mooncast} cite this as a significant bottleneck to scaling SLMs on more data.
It also bottlenecks the speed at which these SLMs can generate tokens.
Recently, new SSL models \cite{cho2024sylber, baade2024syllablelm} have been developed which derive SSL based features at a syllable level resolution, rather than a frame level resolution. 
These models use SSL to naturally segment and drive higher-level features correlated to syllables naturally from the data.
In particular, Sylber \cite{cho2024sylber} was able to achieve a representation at 4.27 Hz while showing some competitive results on some SLU tasks.
However, due to this reduction in sampling rate, it is unclear how language models trained with syllabic speech tokenization scale compared to more traditional SSL based counterparts.

In this work, we present the first systematic study of syllabic tokenization for spoken language modeling.
We evaluate and compare SLM trained with Sylber-based tokenization with Hubert-based tokenization at different data scales in order to see how syllabic tokenization stacks up against frame-wise tokenization.

\section{Related Work}

Spoken Language Models (SLMs) have seen an increase in popularity and research as researchers in academia and industry pretrain models on more and more audio data, following the scaling trends in the LLM space.
Early work in SLMs emerged from the Generative Spoken Language Model (GSLM) \cite{gslm} line of work which were among the first to quantize SSL models and use them as tokens for speech models. 
Later, TWIST \cite{hassid2023textually} was one of the first to use text-pretrained models to initialize SLMs, taking advantage of the abundance of strong text-based LLMs being released.
Currently, practically all modern SLMs trained on millions of hours of data \cite{defossez2024moshi, ju2025mooncast, ding2025kimi} use a text pretrained model to initialize, or include text as an extra modality during training.

Scaling laws provide a framework to derive the optimal way to allocate compute to model size and dataset size to get the best performance, such that future researchers can use these parameters as guidelines to train future models.
Recent work \cite{cuervo2024scaling, maimon2025scaling} has applied these scaling laws to SLMs and found that their convergence speed seems to be 3 orders of magnitude slower than with LLMs.
Notably, they found that Unigram tokenization of Hubert tokens scaled worse downstream, indicating a fault in using BPE to reduce context length.

\section{Experimental Setup}

In this work, we use the Slamkit framework \cite{maimon2025slamming} in order to compare Sylber and Hubert based tokenization for SLMs.

\subsection{Model Architectures}

To build syllabic speech tokenizers, we largely follow the previous study \cite{cho2024sylber}, using the official checkpoint of Sylber to extract segment-averaged embeddings, which we then tokenize with k-means clustering. We used the LibriSpeech data with varying numbers of clusters (5000, 10,000, 20,000 and 40,000).

For the vocoder, we trained a Conditional Flow-Matching Model (CFM) \cite{lipman2022flow}.
Since Sylber units remove the silence between syllables, we trained a CFM model to predict both the duration of the unit itself and the duration of the silence before the unit.
From there, we pass these units with the duration and silence information as well as a speaker embedding derived from the L0 layer of WavLM-base-plus \cite{chen2022wavlm} to a separate CFM module, which decodes into mel-spectrogram.
Both CFM models were trained on the LibriTTS \cite{zen2019libritts} and EXPRESSO \cite{nguyen2023expresso} datasets.
We use an off the shelf vocoder from SpeechBrain \cite{speechbrain_v1} to decode the mel-spectrogram to 16kHz audio.

For our baseline, we use the Hubert tokenizer and vocoder provided in \cite{maimon2025slamming}.
This tokenizer has a vocab size of 500 and we deduplicated the units, which reduced the sampling rate by half, to 25Hz.

In our study, we used two base models, OPT-125M \cite{zhang2022opt} and Qwen2.5-0.5B \cite{qwen2.5}, and we use a TWIST-style \cite{hassid2023textually} initialization for our models.

\subsection{Datasets}

For each of the tokenizers, we train 3 separate models. 
We first pre-train with LibriSpeech \cite{panayotov2015librispeech}, then pre-train with Librispeech and Librilight \cite{kahn2020libri} and finally pre-train with Librispeech, Librilight and Spoken TinyStories \cite{maimon2025slamming}.
This way, we can map the improvement of the model as we add more and more pretraining data to the mix.
We use this particular mixture because it was the best performing mixture found in \cite{maimon2025slamming}.
All of the models are trained for 1 epoch and we use the hyperparameters derived from \cite{maimon2025slamming}.

\subsection{Evaluation Metrics}
\label{sec:metrics}

We used 4 different metrics to evaluate our model \cite{maimon2025slamming}.
\textbf{sBLIMP} \cite{dunbar2021zero} was used to evaluate syntactic understanding, which is a spoken version of the BLIMP dataset, which has the model differentiate between grammatical and ungrammatical pairs of spoken sentences. In this study, we use the dev set.
\textbf{Spoken Story-Cloze (sSC)} \cite{hassid2023textually} is a dataset of spoken sentences from a story where the goal is to distinguish an irrelevant sample from the rest. \textbf{Topic Story-Cloze (tSC)} \cite{hassid2023textually} is a simplified version of sSC where the negative sentence is from a completely different topic.
We measure \textbf{Generation Perplexity (GenPPL)} as defined by \cite{maimon2025slamming}.
Here, we provide the SLM with short speech prompts and generate speech tokens. 
A vocoder is used to convert the tokens into speech, which is transcribed and evaluated using an LLM.
The ASR model used is Whisper-large-v3-turbo \cite{radford2023robust} and the LLM that we use to measure perplexity is Llama-3.2-1B. \cite{grattafiori2024llama}
We take 1000 random correct samples from sSC and use the first 3 seconds of audio as input prompts to generate the continuations.

One caveat here is that Sylber tokens have a 5x coarser representation than the deduplicated Hubert tokens.
Given tokenized sequences of equal length, the Sylber based continuations would be 5x longer.
To try and fairly generate speech of a similar length, we set the max length to 150 for Hubert \cite{maimon2025slamming}, and 30 for Sylber.

\section{Results and Analysis}

\begin{table}
  \centering
  \caption{Training tokens processed by each tokenizer.}
  \vspace{0.5em}
  \label{tab:token_stats}
  \begin{tabular}{lrr}
    \toprule
    \textbf{Dataset} & \textbf{Hubert} & \textbf{Sylber} \\
    \midrule
    LibriSpeech               & 66.4 M  & 13.1 M \\
    LibriLight                & 3.75 B  & 0.76 B \\
    sTinyStories                       & 2.23 B  & 0.46 B \\
    \midrule
    \textbf{Total}            & 6.04 B  & 1.24 B \\
    \bottomrule
  \end{tabular}
  \vspace{-1em}
\end{table}

\begin{table*}[htbp]
  \centering
  \vspace{-1em}
  \caption{Tokenizer comparison across datasets for \textbf{Qwen 2.5-0.5B} and \textbf{OPT-125 M}.  
           Higher $\uparrow$ is better except for GenPPL $\downarrow$}
  \label{tab:tokenizer_eval}
  \begin{tabular}{lllrrrrrrrr}
    \toprule
    \multicolumn{3}{c}{\textbf{Tokenizer}} &
    \multicolumn{4}{c}{\textbf{Qwen 2.5-0.5B}} &
    \multicolumn{4}{c}{\textbf{OPT-125M}} \\
    \cmidrule(r){1-3}\cmidrule(l){4-7}\cmidrule(l){8-11}
    Model & Vocab & Tokens &
    sBLIMP $\uparrow$ & sSC $\uparrow$ & tSC $\uparrow$ & GenPPL $\downarrow$ &
    sBLIMP $\uparrow$ & sSC $\uparrow$ & tSC $\uparrow$ & GenPPL $\downarrow$ \\
    \midrule
    \multicolumn{11}{l}{\textbf{LibriSpeech}}\\
    Hubert  & 500  & 66.4M & 53.02 & \textbf{53.77} & \textbf{64.78} & \textbf{237.19} & 51.95 & 54.68 & 65.26 & 279.45 \\
    Sylber  & 5k   & 13.1M & 56.49 & 51.79 & 55.26 & 348.64 & 55.86 & 51.42 & 53.77 & 365.53 \\
            & 10k  & 13.1M & \textbf{57.48} & 50.35 & 54.36 & 356.75 & 56.49 & 50.19 & 53.29 & 359.54 \\
            & 20k  & 13.1M & 55.34 & 52.16 & 52.11 & 316.03 & 55.09 & 50.77 & 52.38 & 369.20 \\
            & 40k  & 13.1M & 55.01 & 52.81 & 51.04 & 386.87 & 54.95 & 52.06 & 51.04 & 377.71 \\
    \midrule
    \multicolumn{11}{l}{\textbf{LibriSpeech + LibriLight}}\\
    Hubert  & 500  & 3.81B & 56.84 & \textbf{53.39} & \textbf{71.03} & \textbf{151.00} & 56.13 & 54.41 & 70.02 & 167.58 \\
    Sylber  & 5k   & 774.9M& 60.56 & 51.79 & 69.91 & 210.92 & 59.55 & 50.03 & 68.09 & 212.47 \\
            & 10k  & 774.9M& \textbf{61.34} & 51.20 & 69.48 & 204.92 & 60.06 & 51.84 & 68.84 & 211.91 \\
            & 20k  & 774.9M& 60.65 & 51.90 & 69.96 & 207.11 & 60.20 & 52.59 & 68.15 & 203.16 \\
            & 40k  & 774.9M& 61.20 & 52.38 & 67.93 & 211.90 & 60.22 & 52.27 & 67.50 & 221.73 \\
    \midrule
    \multicolumn{11}{l}{\textbf{LibriSpeech + LibriLight + STS}}\\
    Hubert  & 500  & 6.04B & 56.95 & 57.30 & 79.64 &  \textbf{85.90} & 56.07 & 56.97 & 76.27 & 107.80 \\
    Sylber  & 5k   & 1.24B & 60.54 & 57.67 & 79.58 & 168.81 & 59.57 & 55.85 & 77.28 & 184.50 \\
            & 10k  & 1.24B & 60.80 & 57.51 & 78.41 & 177.69 & 59.93 & 56.23 & 76.96 & 177.12 \\
            & 20k  & 1.24B & 60.57 & \textbf{58.90} & \textbf{80.17} & 183.08 & 60.42 & 56.12 & 79.05 & 181.81 \\
            & 40k  & 1.24B & \textbf{60.83} & 57.30 & 78.46 & 187.17 & 60.01 & 55.96 & 76.64 & 172.68 \\
    \bottomrule
  \end{tabular}
  \vspace{-0.5em}
\end{table*}

\begin{figure*}[htbp]
  \centering
  \vspace{-0.5em}
  \includegraphics[width=\textwidth]{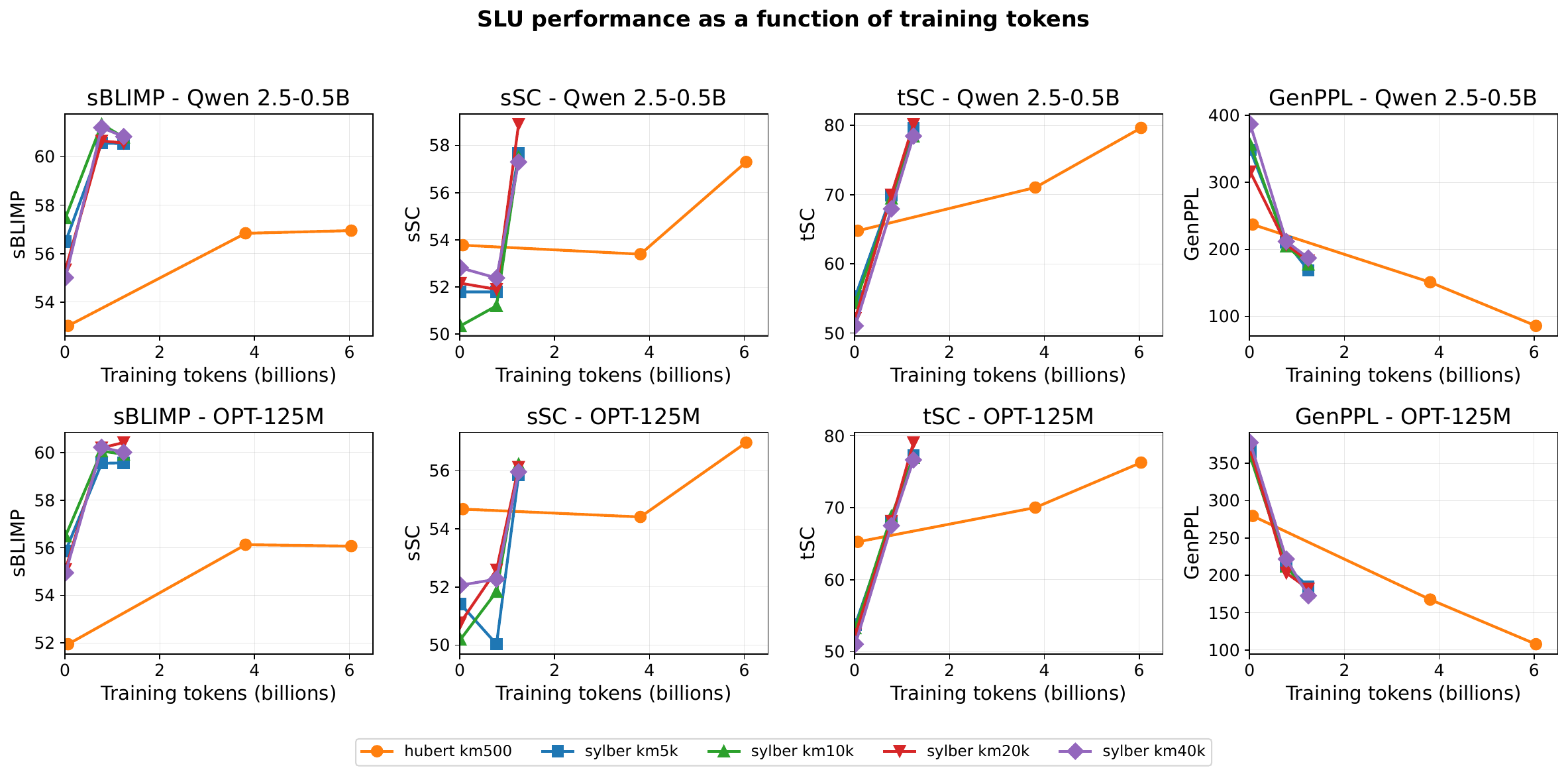}
  \caption{%
    \textbf{Model performance as a function of training-token budget.}
    Each panel shows one evaluation metric and one base model size.
    Higher values are better for all metrics except GenPPL, where lower is
    better.
  }
  \label{fig:opt_token_scaling}
  \vspace{-1em}
\end{figure*}

The results of our experiments are shown in Table \ref{tab:tokenizer_eval}.
The table is broken into 3 parts, for each of the pretraining mixes that we decided to use.
The first 3 columns are for the tokenizer, which show which SSL model it is based on, as well as the vocabulary size, and the total number of tokens in the training data.
As we can see, the number of tokens overall is about 5x less for the Sylber based tokenizer compared to Hubert across the board.
You can also see this in Table \ref{tab:token_stats} which shows the statistics for the number of tokens for each of the datasets after tokenization.
The 8 columns on the right are split into 2 sections, the first for the larger Qwen2.5-0.5B and the second for the smaller OPT-125M based model.
Each section has 4 columns corresponding to the 4 different metrics detailed in Section \ref{sec:metrics}.

In Figure \ref{fig:tsc_comparison}, we plot the performance of the Sylber-20k models on tSC against training compute against other baselines in \cite{maimon2025slamming}. 
These models perform well at their model size, notably performing on par with Slam (-DPO) with significantly fewer training flops.
The Slam model shown above Sylber Qwen-0.5B shows better performance, but note that this model uses additional preference training we did not use.

\subsection{Comparing Hubert and Sylber}

In order to make a comparison between Hubert and Sylber, we plotted the trends for these experiments in Figure \ref{fig:opt_token_scaling}, which shows how the tokenizers compare with each other.

Sylber always performs better than Hubert on sBLIMP across all data points.
This is interesting, as the Sylber model sees 5x less data than the Hubert based model and outperforms it on all metrics.
For sSC, the performance charts show that the Sylber model trained on only LibriLight and Librispeech performs worse than Hubert overall, but significantly outperforms Hubert once STS is introduced.
On the other hand, for tSC, the trend line for Sylber is more linear, and matches the performance of Hubert when trained on at least LibriSpeech and LibriLight. 
For GenPPL, the trendline for the Sylber model is steeper compared to the Hubert model, suggesting that Sylber-based models may converge quicker in that aspect.

The 5x reduction in context length can also be seen in the training time reduction.
On an 8xA100-80GB NVIDIA DGX system, the final Hubert-based model trained on all 3 datasets takes 8.5 hours to complete while the Sylber KM20000 based model only takes 3 hours.

\vspace{-1em}
\subsection{Vocabulary Size}

For the Sylber-based SLM, we used 4 different vocab sizes to see how this would affect performance.
For the most part, increasing the vocab size does not appear to affect performance.
Overall, 20,000 appears to be the best vocabulary size overall, with performance most consistently at the top. 
As noted by \cite{cho2024sylber}, the naive k-means might be suboptimal to discretize syllable space given the combinatorial nature of it.

\subsection{Correlations between Datasets and Benchmarks}

Due to the way that we pretrained our models, it also shows us some correlations between the pretraining mixture and performance. Notably, sSC scores improve across the board when adding sTinyStories to the training mix, which is an insight that \cite{maimon2025slamming} also observed for Hubert based models.
On the other hand, introducing sTinyStories into the pretraining mix has a muted effect on sBLIMP where it sometimes decreases performance, such as with the Sylber-based OPT models as shown in Figure \ref{fig:opt_token_scaling}.
These trends show that the insights we had from Hubert may also apply to Sylber based models.

\vspace{-1em}
\section{Conclusion}
\vspace{-0.5em}

In conclusion, we conducted a study comparing Sylber-based SLMs to Hubert-based SLMs.
We found that Sylber-based models can match or surpass Hubert-based models on sBLIMP, sSC, and tSC with a 5x reduction in training tokens, showing that Sylber-based SLMs are a viable and efficient alternative to frame-level SLMs.

\vspace{-1em}
\section{Acknowledgments}
\vspace{-1em}

This work was supported in part by the NVIDIA Academic Grant Program award. 
Special thanks to Shang-Wen Li and Ching-Feng Yeh for their advice and input.

\section{Compliance with Ethical Standards}
\vspace{-0.5em}
This is a machine learning study for which no ethical approval was required.
\vspace{-1em}

\bibliographystyle{IEEEbib}
\bibliography{refs}

\end{document}